\newif\ificraofficial
\IfFileExists{ieeeconf.cls}{%
  \icraofficialtrue
  \documentclass[letterpaper,10pt,conference]{ieeeconf}
}{%
  \icraofficialfalse
  \documentclass[letterpaper,10pt,conference]{IEEEtran}
}

\ificraofficial
  \IEEEoverridecommandlockouts
\else
  \IEEEoverridecommandlockouts
\fi

\usepackage{amsmath,amssymb}
\usepackage{graphicx}
\usepackage{booktabs}
\usepackage{url}

\usepackage{hyperref}
\usepackage[
    backend=biber,
    style=ieee,
    sorting=none,
    natbib=true,
    doi=false,
    isbn=false,
    url=false,
    eprint=true,
    maxcitenames=1,
    mincitenames=1,
    citestyle=numeric-comp
]{biblatex}

\usepackage[printonlyused]{acronym}

\usepackage{siunitx}
\usepackage[all]{nowidow}

\usepackage[table]{xcolor}

\usepackage{tikz}
\usetikzlibrary{arrows}
\usetikzlibrary{arrows.meta}
\usepackage{makecell}
\usetikzlibrary{positioning}
\usetikzlibrary{decorations.pathreplacing,calligraphy}
\usepackage[normalem]{ulem}

\usepackage{lipsum}

\usepackage{hhline}

\usepackage{xspace} %smart handling of space in commands
\newcommand{\ie}{i.e.,\xspace}

\usepackage{epstopdf}

\usepackage{import}

\usepackage{array}
\usepackage{longtable}
\newcolumntype{L}{>{\raggedright\arraybackslash}p{0.13\textwidth}}
\newcolumntype{C}{>{\centering\arraybackslash}p{0.10\textwidth}}
\newcolumntype{Y}{>{\centering\arraybackslash}m{2.2cm}}
\newcolumntype{M}{>{\centering\arraybackslash}p{0.065\textwidth}}

\usepackage{tabularx}
\usepackage{multirow, multicol}

\newlength{\Oldarrayrulewidth}

\newcolumntype{?}[1]{!{\vrule width #1}}

\usepackage{amsfonts,amscd}

\usepackage{bm}

\usepackage{balance}
\usepackage[T1]{fontenc}

\usepackage{cancel}

\newcommand{\bbm}{\begin{bmatrix}}
\newcommand{\ebm}{\end{bmatrix}}

\newcommand{\ignore}[1]{}

\newcommand{\bma}[1]{\left[\begin{array}{#1}}
\newcommand{\ema}{\end{array}\right]}

\DeclareMathAlphabet{\mbf}{OT1}{ptm}{b}{n}

\def\fdotb{{\raisebox{-0.6ex}{ \kern0.2ex\raisebox{0.8ex}{\tiny $\hspace*{-1ex}\circ$}}}}
\def\fddotb{{\raisebox{-0.6ex}{ \kern0.2ex\raisebox{0.8ex}{\tiny $\hspace*{-1ex}\circ\circ$}}}}
\newcommand{\utimes}{ {\raisebox{-0.6ex}{ \kern-1.0ex\raisebox{0.6ex}{ \small $\mathsf{v}$}}} } % 
\newcommand{\beq}{\begin{equation}}
\newcommand{\eeq}{\end{equation}}
\newcommand{\bdis}{\begin{displaymath}}
\newcommand{\edis}{\end{displaymath}}
\newcommand{\beqarray}{\begin{eqnarray}}
\newcommand{\eeqarray}{\end{eqnarray}}
\newcommand{\beqarraynn}{\begin{eqnarray*}}
\newcommand{\eeqarraynn}{\end{eqnarray*}}

\DeclareMathAlphabet{\mbf}{OT1}{ptm}{b}{n}

\newcommand{\papertitle}{Dr-LiSA: Direct Radar-Lidar Scan Alignment for $SE(3)$ Localization}
\ificraofficial
  \title{\LARGE \bf \papertitle}
\else
  \title{\papertitle}
\fi

\author{
    Alex Zhang$^{1}$,
    Daniil Lisus$^{1}$,
    Cedric Le Gentil$^{2}$,
    Timothy D. Barfoot$^{1}$
    \thanks{$^{1}$Robotics Institute, University of Toronto, Canada}
    \thanks{$^{2}$Mobile Robotics Lab, ETH Z\"{u}rich, Switzerland}
    \thanks{Corresponding: \texttt{alex.zhang@robotics.utias.utoronto.ca}}
}

\begin{document}

\maketitle
\thispagestyle{empty}
\pagestyle{empty}

\begin{abstract}
This paper introduces Dr-LiSA, a first-of-its-kind direct method for localizing 2D spinning radar intensity measurements in $SE(3)$ against 3D lidar maps. Radar-lidar localization combines the complementary strengths of the two sensing modalities: radar is robust to adverse weather and precipitation, while lidar provides high-fidelity 3D maps in favourable conditions. However, existing radar-lidar localization methods are restricted to planar $SE(2)$ localization and have generally fallen short of the accuracy achieved by lidar-lidar and even radar-radar systems. A key challenge is the substantial sensing-modality gap between radar and lidar, which observe and represent scene structure in fundamentally different ways. Dr-LiSA bridges this gap using a learned forward model that predicts radar measurements from a lidar submap at a candidate pose, enabling direct photometric alignment of predicted and observed radar scans in $SE(3)$. Dr-LiSA outperforms prior radar-lidar approaches in $SE(2)$ while achieving planar accuracy competitive with state-of-the-art radar-radar localization across more than \SI{90}{\km} of on-road data.
\end{abstract}

\section{Introduction}
\label{sec:introduction}

Accurate and reliable localization is a fundamental requirement for autonomous driving. Global Navigation Satellite Systems (GNSS) coupled with inertial sensing can provide globally referenced pose estimates, but autonomous vehicles often require localization relative to mapped environmental structure. Moreover, the accuracy and availability of GNSS can degrade in environments such as urban canyons, tunnels, and other signal-denied regions~\cite{s26165316}. Map-based localization therefore remains an important component of autonomous driving systems, where real-time exteroceptive sensor observations are registered against a previously constructed map to obtain precise vehicle pose estimates.

Lidar is particularly well-suited to map-based localization due to its accurate and dense geometric measurements, and lidar-lidar localization methods (\ie using real-time lidar scans to localize against preconstructed lidar maps) have achieved centimetre-level localization accuracy. However, lidar measurements can degrade in adverse weather due to attenuation and backscatter from precipitation. Radar offers a complementary sensing modality: its longer wavelength makes it comparatively robust to small particles in the air such as precipitation, but its measurements are inherently noisier and lower resolution. This motivates radar-lidar localization, where a weather-robust radar sensor is localized against a high-quality 3D lidar map constructed under favorable conditions. Such an approach also enables existing lidar maps to be reused for radar localization, avoiding the need to construct and maintain a separate radar map.

Despite these practical advantages, existing radar-lidar methods have yet to match the accuracy of lidar-lidar localization and generally also trail state-of-the-art (SOTA) radar-radar approaches~\cite{are_we_ready_for}. A central difficulty is the large sensing-modality gap between radar and lidar: the two sensors respond to different physical properties of the environment and therefore do not produce directly comparable representations. Moreover, existing radar-lidar approaches are typically formulated in $SE(2)$. Because spinning radars produce 2D range-azimuth images, prior methods commonly reduce the 3D lidar map to a planar or bird's-eye-view (BEV) representation before registration. While this simplifies cross-modal matching, it discards vertical scene structure and fundamentally prevents estimation of the full $SE(3)$ vehicle pose.

\begin{figure}[t]
    \centering
    \includegraphics[width=\columnwidth]{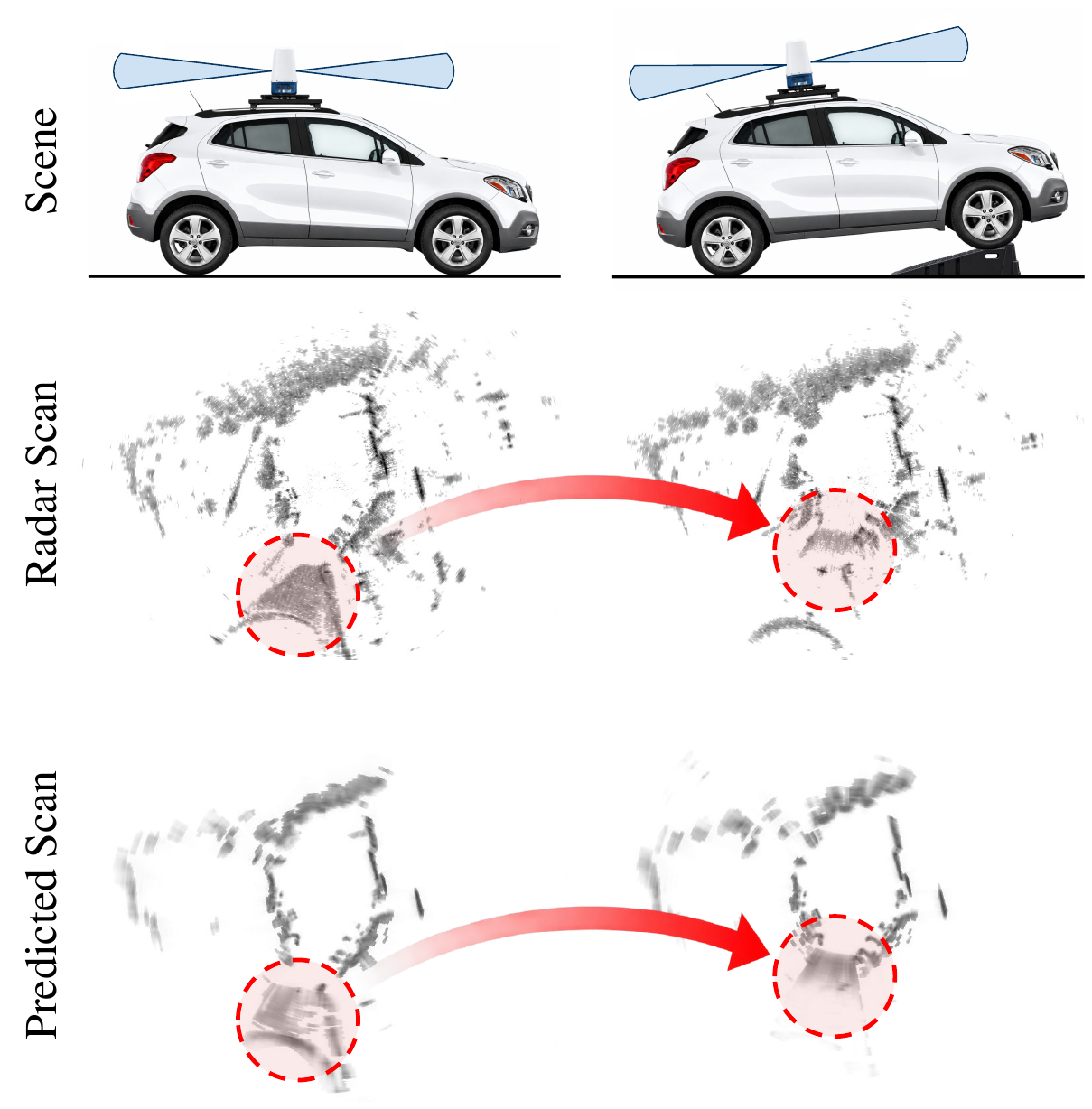}
    \caption{Real-world example of 3D information in spinning-radar measurements. Two poses with similar $SE(2)$ components but different pitch produce distinct ground-plane returns in the measured and predicted radar scans, illustrating the non-planar information exploited by Dr-LiSA.}
    \label{fig:concept-intro}
\end{figure}

Although the radar output is fundamentally 2D, the underlying sensing process is not. A spinning radar illuminates the environment with a 3D beam; thus, the measured intensities depend on how objects in the scene interact with that beam. Consequently, changes in radar height, roll, and pitch can alter the observed 2D intensity image even though these dimensions are not explicitly represented. This raises the natural question of whether 2D spinning-radar images encode sufficient information for 3D localization. In this work, we exploit this implicit dependence on 3D geometry rather than collapsing the lidar map to two dimensions.

We introduce a Direct Radar-Lidar Scan Alignment (Dr-LiSA) framework that estimates the radar pose in $SE(3)$ against a 3D lidar map. Given a candidate $SE(3)$ radar pose, Dr-LiSA extracts the locally relevant 3D geometry from the lidar submap and uses a learned forward model to predict the radar intensity measurement that would be observed from that pose. Localization is then formulated as direct alignment: the candidate pose is optimized by minimizing the pixel-wise photometric error between the predicted and measured radar scans. Unlike feature-based approaches, which reduce measurements to a sparse set of selected landmarks or descriptors, the direct formulation exploits the dense intensity information available in the spinning-radar image. More importantly, because the prediction explicitly depends on the radar's full 3D pose relative to the lidar map, variations in all six degrees of freedom can influence the alignment objective.

This 3D awareness is particularly valuable when the vehicle experiences appreciable roll or pitch, or when conventional planar localization has little horizontal structure to exploit. Fig.~\ref{fig:concept-intro} illustrates two nearly identical $SE(2)$ poses that differ primarily in pitch, revealing a clear change in the observed ground-plane returns. A planar representation may discard or suppress this feature, whereas Dr-LiSA exploits it through its 3D forward model. We show that exploiting 3D scene structure improves localization robustness in feature-sparse, non-planar sections that are challenging for traditional planar methods.

The key contributions of this paper are as follows:
\begin{itemize}
    \item Dr-LiSA: the first $SE(3)$ localization framework using spinning radar, and the first framework for direct radar-lidar localization. We validate the approach on more than \SI{90}{\km} of real-world driving across diverse and challenging environments.

    \item A 3D-aware learned lidar-to-radar model that predicts radar intensity from local 3D geometry.

    \item A demonstration that 3D localization improves robustness in feature-sparse, non-planar environments.
\end{itemize}

\begin{figure*}[t]
    \centering
    \includegraphics[width=\textwidth]{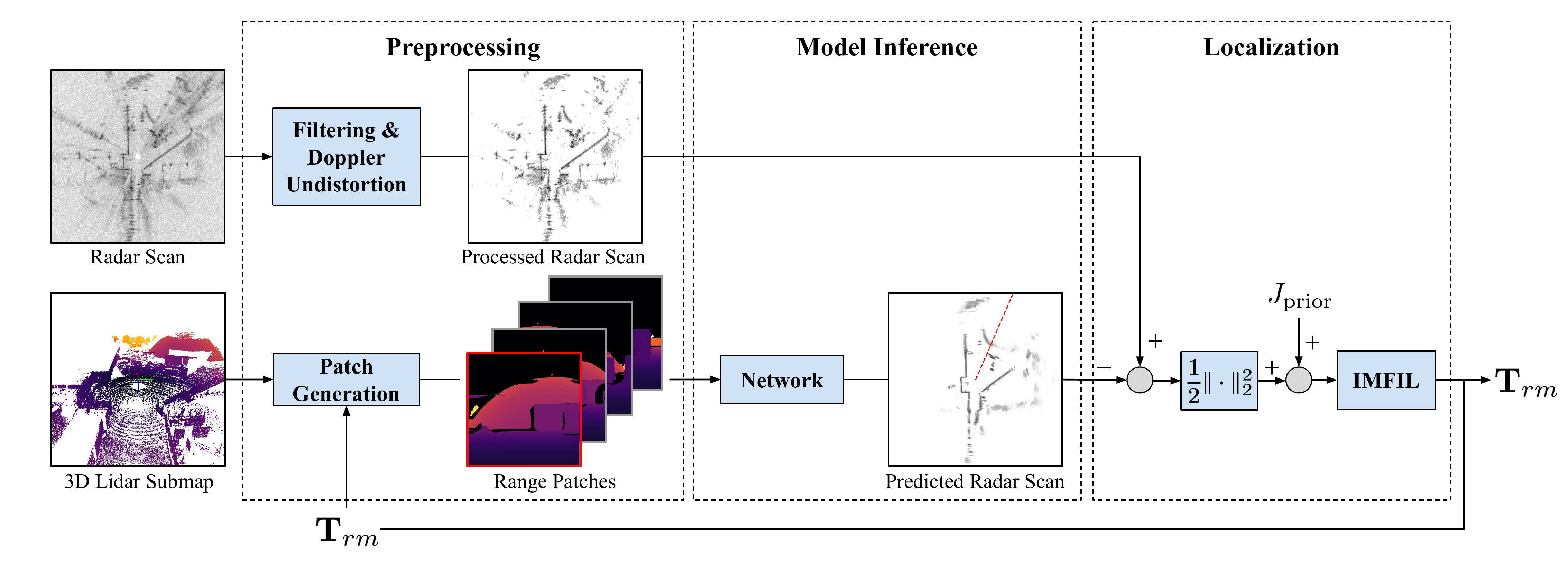}
    \caption{
    Overview of the Dr-LiSA localization pipeline. A processed radar scan is compared against a radar scan predicted from pose-dependent lidar range patches. The photometric error and pose prior are minimized with IMFIL to iteratively refine $\mathbf{T}_{rm}$, the radar pose relative to the lidar submap. Radar scans are shown in Cartesian coordinates for visualization, although alignment is performed in polar coordinates; the red-bordered patch corresponds to the red dashed azimuth in the predicted radar scan.
    }
    \label{fig:block-diagram}
\end{figure*}

\section{Related Work}
\label{sec:related_work}

Radar has seen renewed interest in robotics due to advances in sensing hardware and radar-based state estimation~\cite{a_new_wave_radar,abu2023radar,venon2022millimeter}. Early spinning-radar odometry and localization methods commonly operate on sparse representations extracted from raw range-azimuth intensity images. \citet{Cen_filtering}, for example, estimate ego-motion using Iterative Closest Point (ICP) registration between extracted radar points. Learning has also been used to improve these representations: \citet{barnes2019masking} learn a radar mask for correlative scan matching; Barnes and Posner~\cite{under_the_radar} learn repeatable keypoints, descriptors, and detection scores; and \citet{burnett_rss21} demonstrate self-supervised radar feature learning without ground-truth supervision. Most radar state-estimation research has focused primarily on odometry and simultaneous localization and mapping (SLAM), with localization against preconstructed radar maps receiving comparatively less attention.

Point-based localization has nevertheless shown that radar can approach the accuracy required for autonomous driving. \citet{are_we_ready_for} use an ICP-based localization approach by extracting radar points with a variant of Constant False Alarm Rate (CFAR)~\cite{rohling1983radar}. Localization is then performed through batch optimization against the radar map, achieving translational errors of \SIrange{6}{10}{\cm}. More recently, direct methods have improved performance by operating on dense radar measurements rather than sparse features. Direct Radar Odometry (DRO)~\cite{legentil2025dro} directly optimizes cross-correlation between radar scans and local maps while accounting for motion and Doppler distortion. Dr-BA~\cite{lisus2026drba} extends this paradigm to bundle adjustment and dense radar map construction, with Direct Radar Localization (DRL) on these maps achieving SOTA performance across diverse environments.

Radar has also been used to localize against preconstructed lidar maps. \citet{Park_Kim_Kim_2019} demonstrate an early multimodal registration framework for indoor disaster environments, recursively compensating for radar motion distortion and geometrically matching live radar scans to a prior lidar map when smoke degraded lidar sensing. Subsequent work has increasingly addressed the radar-lidar modality gap through learned representations. \citet{radar_on_lidar} use a generative model to transform radar measurements toward a lidar-like representation for Monte Carlo localization~\cite{mcl}, while \citet{RaLL} map radar and lidar into a shared embedding space and performs cross-correlation for localization. The authors of \cite{RoLM} instead group radar and lidar points in Cartesian and polar space, obtain an initial alignment using a density metric, and refine it with ICP. \citet{Yin_Xu_Wang_Xiong_2021} similarly learn a shared representation for heterogeneous place recognition, but do not evaluate metric localization accuracy. \citet{wang2020l2rgan} synthesize radar images from lidar BEV inputs using a conditional generative model. Although conceptually similar to our lidar-to-radar forward model, it is limited to planar representations and is not evaluated for downstream localization.

Although promising, these earlier radar-lidar approaches generally report global localization errors on the order of \SI{1}{\m} or greater, well above the centimeter-level accuracy targeted for safe autonomous vehicle operation~\cite{reid2019localization}. More recent methods have achieved substantially higher accuracy through geometric registration. \citet{are_we_ready_for} introduce Radar-Lidar Teach and Repeat (RLT\&R), which extracts radar point clouds using CFAR-based detectors and registers them against preconstructed lidar submaps with ICP. This approach achieves radar-lidar localization accuracy on the order of tens of centimeters. However, CFAR-based extraction relies on local measurement statistics and can retain substantial noise. \citet{lisus2025pointing} address this by using a U-Net~\cite{unet} to weight extracted radar points, improving isolated ICP trials but leaving full localization pipeline integration as future work. Overall, existing radar-lidar localization methods remain predominantly point-based and restricted to $SE(2)$ pose estimation.

\section{Methodology}
\label{sec:methodology}

Although spinning-radar intensity images are two-dimensional, their measured intensities depend on interactions between the radar beam and surrounding 3D scene structure. Dr-LiSA exploits this dependence to enable localization in $SE(3)$ from 2D radar measurements. We first use the Teach and Repeat (T\&R) framework~\cite{are_we_ready_for, VTR} to accumulate sequential lidar measurements offline into a series of connected local submaps for reuse during subsequent repeat traversals. Radar-lidar localization is then formulated as a direct photometric scan-alignment problem, in which real-time radar measurements are matched against predictions synthesized from 3D lidar submaps. Given an initial pose, a learned forward model generates the corresponding radar scan, and the pose estimate is optimized by minimizing the photometric error between the predicted and observed measurements. The overall pipeline, illustrated in Fig.~\ref{fig:block-diagram}, consists of three stages: Preprocessing, Model Inference, and Localization.

\subsection{Preprocessing}
\label{sec:preprocessing}

\textit{1) Lidar:} Rather than providing the full 3D lidar submap to the forward model, we represent the local scene at each radar azimuth using a compact, pose-dependent spherical range-image patch. For a given radar pose within the active submap, each patch is centered on the corresponding azimuth direction and spans $\pm3^\circ$ in both azimuth and elevation with an angular resolution of $0.1^\circ$, yielding a $61 \times 61$ range image. Each pixel stores the distance to the nearest visible surface along its corresponding azimuth-elevation ray. This representation retains only the local geometry relevant to a particular radar azimuth while substantially reducing the model input dimensionality. Because the patch is rendered from the radar pose associated with each azimuth, it also allows intra-scan motion to be incorporated directly into the model input.

To efficiently generate these patches for arbitrary candidate poses, each lidar submap is reconstructed offline as a continuous surface mesh using NKSR~\cite{huang2023nksr} and cached for runtime use. Range patches are then rendered with the NVIDIA OptiX ray-tracing engine~\cite{parker2010optix} and restricted to the \SIrange{7}{120}{\m} interval, excluding close-range geometry that is particularly susceptible to occlusion.

\textit{2) Radar:} The raw polar radar scans are preprocessed to correct systematic measurement effects and suppress undesirable returns. Doppler-induced distortions are compensated using standard motion correction techniques as described in~\cite{2021_Burnett}. The radar measurements are also restricted to the same \SIrange{7}{120}{\m} interval, corresponding to $N_r$ retained range bins with measurements below \SI{7}{\m} set to zero. Raw radar intensity images contain substantial noise and artefacts that make both model training and direct scan alignment unreliable. We therefore apply the same filtering strategy as~\cite{lisus2025doppler} independently to each azimuth waveform (based on~\cite{Cen_filtering}).

\subsection{Model Inference}
\label{sec:model_inference}

\begin{figure*}[t]
    \centering
    \includegraphics[width=\textwidth]{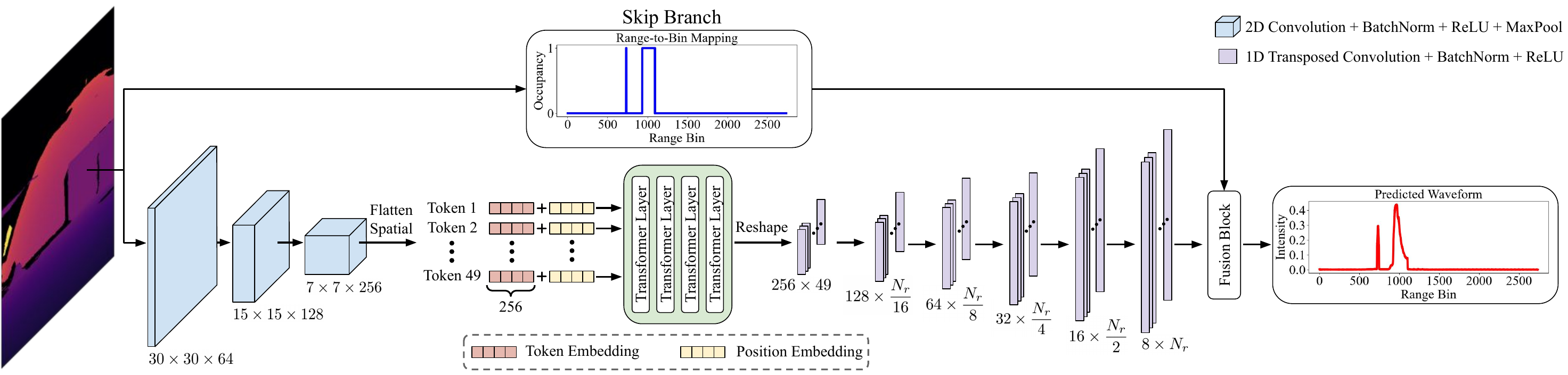}
    \caption{Overview of the model architecture. A range-image patch is passed through an encoder-decoder branch and a geometry-guided skip branch which produces a one-dimensional binary occupancy waveform. Their outputs are fused to produce the predicted 1D radar intensity waveform.}
    \label{fig:model_architecture}
\end{figure*}

Our overall model architecture is shown in Fig.~\ref{fig:model_architecture}. The model takes as input a single lidar range-image patch centered at a given radar azimuth and predicts the corresponding one-dimensional radar intensity waveform. Applying the model independently across all azimuths and stacking the predicted waveforms yields the complete polar radar scan. The architecture is implemented in PyTorch~\cite{pytorch} and consists of two branches: a main encoder-decoder branch and a geometry-guided skip branch.

\textit{1) Encoder-Decoder:} Inspired by Vision Transformer-based architectures~\cite{ranftl2021vision, zheng2021rethinking, carion2020end, dosovitskiy2020image}, the main branch comprises a 2D convolutional neural network (CNN) backbone for local feature extraction, a Transformer encoder for global context modeling, and a 1D transposed-convolution decoder that progressively upsamples to the cropped radar range resolution. The CNN backbone processes the single-channel range image through three convolutional blocks before the resulting feature map is flattened, combined with positional embeddings to preserve spatial identity, and processed by a Transformer encoder. This allows spatial features to interact with geometry throughout the entire input patch, which is useful because the structures contributing most strongly to the radar return are not known a priori. The encoded tokens are then rearranged into channel-first format and upsampled through the decoder.

\textit{2) Skip Branch:} We additionally introduce a geometry-guided skip branch that explicitly incorporates the known metric correspondence between lidar range and radar range bins. Through a range-to-bin mapping, each valid lidar range is assigned directly to its corresponding radar bin using the known radar range resolution, producing a one-dimensional binary occupancy waveform with $N_r$ bins. Each bin is set to one if at least one lidar return maps to it and zero otherwise. This allows the learned components to focus primarily on determining which scene geometry contributes to the radar return and the corresponding intensity response, rather than simultaneously learning where that geometry should appear along the range axis. The resulting geometry-guided waveform is concatenated channel-wise with the eight decoder feature channels, yielding a nine-channel representation at the full radar range resolution. A final fusion block combines the two branches using a 1D convolution with 8 output channels (kernel size 5, padding 2) followed by BatchNorm and ReLU, and a second 1D convolution reducing from 8 channels to a single output channel using the same kernel size and padding. The resulting logits are passed through a sigmoid activation to produce the predicted radar intensity waveform.

\textit{3) Training Loss:} A conventional mean-squared-error (MSE) loss over the complete radar waveform is poorly suited to this cross-modal prediction task. First, filtered radar measurements remain imperfect and can contain artefacts such as multipath returns. Second, radar may observe structures that are absent from the lidar submap (or vice versa), making these measurements impossible to infer from the available model input. Finally, the filtered radar waveform is highly sparse; consequently, an MSE loss evaluated uniformly over all range bins is dominated by zero-intensity regions and can encourage a trivial low-intensity prediction.

We therefore introduce a co-visibility loss that concentrates supervision on range regions supported by both sensing modalities while subsampling the remaining background. For this loss, we reuse the same one-dimensional binary occupancy waveform introduced in the skip branch, obtained by mapping each valid lidar range directly to its corresponding radar range bin. Let $l_k \in \{0,1\}$ denote the resulting lidar occupancy at bin $k$, where $l_k=1$ indicates that at least one lidar return maps to that bin. Also, let $r_k$ denote the measured radar intensity at range bin $k$. Using an intensity threshold $\tau$, we identify the set of co-visible range bins as
\begin{equation}
\mathcal{C}
=
\left\{
k \mid l_k = 1 \;\land\; r_k > \tau
\right\},
\qquad
\tau = 10^{-2}.
\end{equation}
Each co-visible bin is expanded by four bins on either side, forming a collective region of interest $\mathcal{R}$. This local window retains the measured radar structure surrounding each lidar-supported return rather than supervising only a single discrete range bin. To prevent the much larger number of background bins from dominating training, we define the eligible background set as
\begin{equation}
\mathcal{B}
=
\left\{
k \notin \mathcal{R}
\mid
l_k = 0 \;\land\; r_k < \tau
\right\}.
\end{equation}
Following a negative-sampling strategy~\cite{mikolov2013distributed}, we randomly sample a subset of negative bins $\mathcal{N} \subset \mathcal{B}$ such that $\lvert\mathcal{N}\rvert~=~10\lvert \mathcal{C}\rvert$. The resulting training mask $\mathcal{M}=\mathcal{R} \cup \mathcal{N}$ then yields the co-visibility loss as
\begin{equation}
\mathcal{L}_{\mathrm{CV}}
=
\frac{1}{|\mathcal{M}|}
\sum_{k\in\mathcal{M}}
\left(m_k-y_k\right)^2,
\quad
y_k=
\begin{cases}
r_k, & k\in\mathcal{R},\\
0, & k\in\mathcal{N},
\end{cases}
\label{eq:covisibility_loss}
\end{equation}
where $m_k$ represents the model predicted intensity at bin $k$.

\subsection{Localization}
\label{sec:localization}

Following the T\&R topometric localization framework~\cite{are_we_ready_for}, we wish to estimate the pose $\mathbf{T}_{rm} \in SE(3)$ of the radar reference frame $\mathcal{F}_r$ relative to the active lidar submap frame $\mathcal{F}_m$. Here we define $\mathbf{T}_{rm}$ as the radar pose at the center azimuth timestamp of each scan. Rather than independently optimizing the pose at every azimuth, we optimize only this reference pose and use radar odometry to determine the intra-scan motion. Specifically, let $\Delta \mathbf{T}_i~\in~SE(3)$ be the relative transformation from $\mathbf{T}_{rm}$ to the pose $\mathbf{T}_{rm,i}$ associated with azimuth $i$ such that
\begin{equation}
\mathbf{T}_{rm,i}
=
\Delta \mathbf{T}_i \mathbf{T}_{rm}.
\label{eq:azimuth_pose}
\end{equation}
Note that the relative transformations $\Delta \mathbf{T}_i$ are held fixed during localization under the assumption that relative odometry drift over the duration of a single radar scan is negligible. Thus, only $\mathbf{T}_{rm}$ is optimized.

We now use the forward model from the previous section to formulate localization as a direct scan-alignment optimization between the processed radar polar image and a radar polar image predicted from the lidar submap. Let $\tilde{\mathbf{m}}_i~\in~\mathbb{R}^{N_r}$ denote the processed radar intensity waveform at azimuth $i$ of the $N_a$ azimuths in a scan, and let $\mathbf{m}_i(\mathbf{T}_{rm,i})~\in~\mathbb{R}^{N_r}$ denote the corresponding predicted waveform. We can then formulate the direct radar alignment cost as
\begin{equation}
J_{\mathrm{radar}}
=
\frac{1}{2}
\sum_{i=1}^{N_a}
\left\|
\tilde{\mathbf{m}}_i
-
\mathbf{m}_i
\!\left(
\Delta \mathbf{T}_i \mathbf{T}_{rm}
\right)
\right\|_2^2.
\label{eq:radar_alignment_cost}
\end{equation}

For initialization, let $\check{\mathbf T}_{rm}\in SE(3)$ denote the prior radar pose obtained by propagating the previous localization estimate using radar odometry. We incorporate a pose prior to regularize the solution toward the odometric initialization,
\begin{equation}
J_{\mathrm{prior}}
=
\left(
\ln\!\left(
\mathbf{T}_{rm}
\check{\mathbf{T}}_{rm}^{-1}
\right)^{\vee}
\right)^{\mathsf{T}}
\boldsymbol{\Sigma}^{-1}
\left(
\ln\!\left(
\mathbf{T}_{rm}
\check{\mathbf{T}}_{rm}^{-1}
\right)^{\vee}
\right).
\label{eq:pose_prior}
\end{equation}
Here, the $\vee$ operator maps an element of $\mathfrak{se}(3)$ to its vector representation in $\mathbb{R}^{6}$~\cite{barfoot2024state}, and $\boldsymbol{\Sigma}$ is the covariance of the pose prior. We use a diagonal $\boldsymbol{\Sigma}$ with translational standard deviations of $(0.5,\,0.2,\,0.2)\,\si{\m}$ and rotational standard deviations of $(1.0,\,1.0,\,0.05)^\circ$. Thus, the posterior pose estimate is given by
\begin{equation}
\hat{\mathbf{T}}_{rm}
=
\underset{\mathbf{T}_{rm}}{\arg\min}
\left(
J_{\mathrm{radar}}(\mathbf{T}_{rm})
+
J_{\mathrm{prior}}(\mathbf{T}_{rm})
\right).
\label{eq:localization_objective}
\end{equation}
Since the localization objective relies on OptiX ray tracing, analytic gradients with respect to pose are not readily available. Moreover, although the objective generally exhibits a well-defined large-scale basin around the optimum, small-scale oscillations can make local gradient estimates unreliable. We therefore solve \eqref{eq:localization_objective} using the bounded Implicit Filtering (IMFIL) algorithm~\cite{kelley2011implicit}, implemented in Scikit-Quant~\cite{lavrijsen2020classical}. IMFIL uses finite-difference estimates over progressively decreasing sampling scales, allowing it to exploit the coarse structure of the objective while remaining robust to small-scale variations.

\section{Experiments}
\label{sec:experiments}

\subsection{Dataset}
\label{sec:dataset}

\begin{figure}[t]
    \centering
    \includegraphics[width=\linewidth]{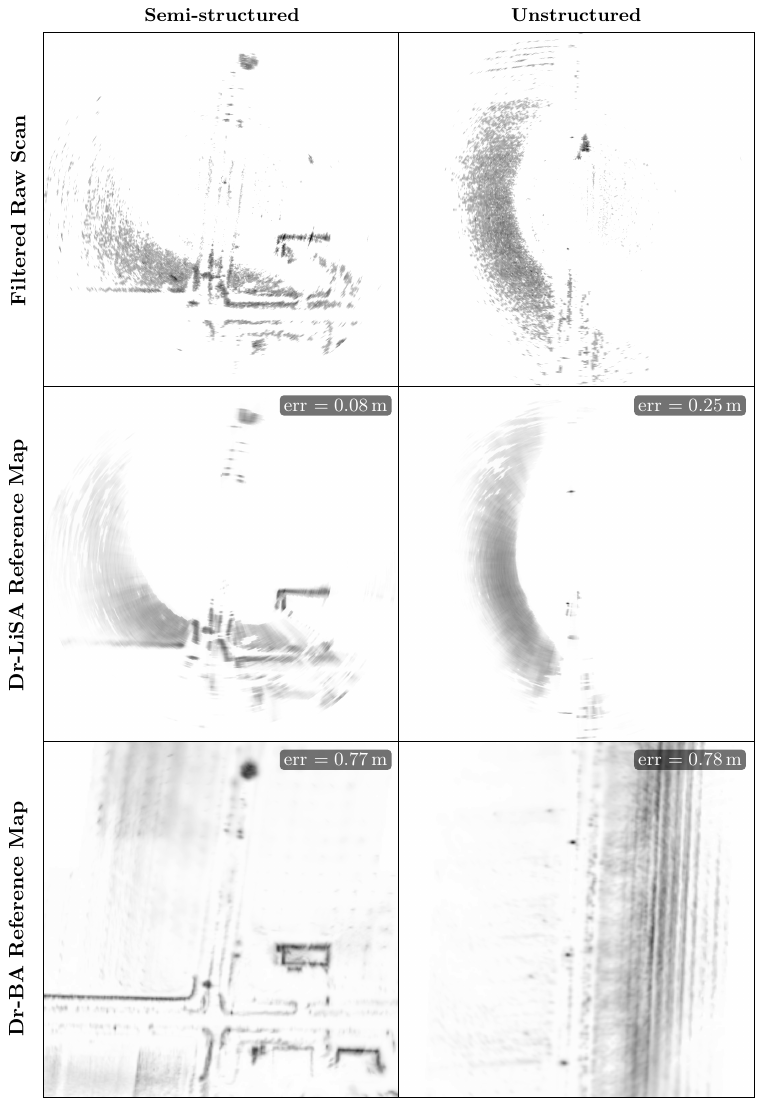}
    \caption{Qualitative comparison on the \texttt{Farm} route. Dr-LiSA more closely aligns with the filtered radar scan than DRL-Dr-BA~\cite{lisus2026drba} in both semi-structured and unstructured regions, with $\mathrm{err}$ denoting planar $SE(2)$ translation error.}
    \label{fig:scan_comparisons}
\end{figure}

\begin{table*}[!t]
    \centering
    \caption{Summary of $SE(3)$ longitudinal, lateral, vertical, roll, pitch, and yaw RMSE localization errors on the Boreas-RT dataset~\cite{lisus_brrt26}. Best radar-based performance is bolded; the lidar-only reference is shown in gray.}
    \footnotesize{
    \begin{tabular*}{\textwidth}{@{\extracolsep{\fill}}llcccccc}
        \toprule
        & & \textbf{Long.} [\si{\m}] & \textbf{Lat.} [\si{\m}] & \textbf{Vert.} [\si{\m}] & \textbf{Roll} [$^\circ$] & \textbf{Pitch} [$^\circ$] & \textbf{Yaw} [$^\circ$] \\
        \midrule
        \multirow{4}{*}{\texttt{Suburbs}}
        & \textbf{RLT\&R} \scriptsize\cite{are_we_ready_for}
        & 0.097 & 0.074 & - & - & - & 0.166 \\
        & \textbf{DRL-Dr-BA} \scriptsize\cite{lisus2026drba}
        & 0.076 & \textbf{0.049} & - & - & - & \textbf{0.061} \\
        & \textbf{Dr-LiSA} \scriptsize(ours)
        & \textbf{0.064} & 0.059 & \textbf{0.172} & \textbf{0.421} & \textbf{0.286} & 0.104 \\
        & \textcolor{gray}{\textbf{2Fast-2Lamaa} \scriptsize\cite{legentil20242fast2lamaa}} & \textcolor{gray}{0.026} & \textcolor{gray}{0.023} & \textcolor{gray}{0.041} & \textcolor{gray}{0.042} & \textcolor{gray}{0.029} & \textcolor{gray}{0.027} \\
        \midrule
        \multirow{4}{*}{\texttt{Industrial}}
        & \textbf{RLT\&R} \scriptsize\cite{are_we_ready_for}
        & 0.097 & 0.066 & - & - & - & 0.150 \\
        & \textbf{DRL-Dr-BA} \scriptsize\cite{lisus2026drba}
        & 0.081 & \textbf{0.062} & - & - & - & \textbf{0.055} \\
        & \textbf{Dr-LiSA} \scriptsize(ours)
        & \textbf{0.067} & 0.065 & \textbf{0.248} & \textbf{0.594} & \textbf{0.348} & 0.109 \\
        & \textcolor{gray}{\textbf{2Fast-2Lamaa} \scriptsize\cite{legentil20242fast2lamaa}} & \textcolor{gray}{0.023} & \textcolor{gray}{0.022} & \textcolor{gray}{0.034} & \textcolor{gray}{0.064} & \textcolor{gray}{0.040} & \textcolor{gray}{0.034} \\
        \midrule
        \multirow{4}{*}{\texttt{Regional}}
        & \textbf{RLT\&R} \scriptsize\cite{are_we_ready_for}
        & 0.115 & 0.097 & - & - & - & 0.235 \\
        & \textbf{DRL-Dr-BA} \scriptsize\cite{lisus2026drba}
        & 0.260 & 0.089 & - & - & - & \textbf{0.078} \\
        & \textbf{Dr-LiSA} \scriptsize(ours)
        & \textbf{0.089} & \textbf{0.087} & \textbf{0.444} & \textbf{0.658} & \textbf{0.363} & 0.131 \\
        & \textcolor{gray}{\textbf{2Fast-2Lamaa} \scriptsize\cite{legentil20242fast2lamaa}} & \textcolor{gray}{0.034} & \textcolor{gray}{0.035} & \textcolor{gray}{0.036} & \textcolor{gray}{0.038} & \textcolor{gray}{0.030} & \textcolor{gray}{0.030} \\
        \midrule
        \multirow{4}{*}{\texttt{Farm}}
        & \textbf{RLT\&R} \scriptsize\cite{are_we_ready_for}
        & -$^3$ & -$^3$ & - & - & - & -$^3$ \\
        & \textbf{DRL-Dr-BA} \scriptsize\cite{lisus2026drba}
        & 0.404 & 0.595 & - & - & - & 0.285 \\
        & \textbf{Dr-LiSA} \scriptsize(ours)
        & \textbf{0.175} & \textbf{0.176} & \textbf{0.990} & \textbf{0.870} & \textbf{0.598} & \textbf{0.200} \\
        & \textcolor{gray}{\textbf{2Fast-2Lamaa} \scriptsize\cite{legentil20242fast2lamaa}} & \textcolor{gray}{0.046} & \textcolor{gray}{0.041} & \textcolor{gray}{0.064} & \textcolor{gray}{0.053} & \textcolor{gray}{0.037} & \textcolor{gray}{0.034} \\
        \bottomrule
        \multicolumn{8}{l}{\scriptsize Superscript indicates number of failed sequences.}
    \end{tabular*}}
    \label{tab:localization_boreas}
\end{table*}

We train and evaluate Dr-LiSA on a subset of the Boreas Road Trip (Boreas-RT) dataset~\cite{lisus_brrt26}, which provides synchronized measurements from a Velodyne Alpha-Prime lidar, a Navtech RAS6 spinning radar, a Silicon Sensing DMU41 6-DoF IMU, and an Applanix RTK-GNSS/INS system used for ground-truth pose estimation. The radar produces 400-azimuth scans with a range resolution of $r_r~=~\SI{0.0438}{\m}$ and a maximum sensing range of approximately \SI{300}{\m}, although only the first \SI{120}{\m} are used for Dr-LiSA.

Our evaluation uses 15 Boreas-RT sequences spanning four distinct route types. For each route, one sequence is used to construct the lidar submaps, for a total mapping distance of \SI{24.1}{\km}, while the remaining 11 sequences are used for localization evaluation. These localization sequences comprise \SI{90.9}{\km} of driving: \texttt{Suburbs} (3), \texttt{Industrial} (3), \texttt{Regional} (2), and \texttt{Farm} (3). The corresponding route lengths are \SI{7.9}{\km}, \SI{5.4}{\km}, \SI{9.3}{\km}, and \SI{10.8}{\km}, respectively. The \texttt{Suburbs} route consists primarily of residential streets and includes traversal of the same roads in opposing directions. The \texttt{Industrial} route passes through a commercial and industrial area with large building structures, with several sequences collected during snowfall. The \texttt{Regional} route contains highway driving with long stretches of repetitive geometric structure. Finally, the \texttt{Farm} route traverses a challenging rural environment with very few buildings and greater road-induced roll and pitch variation.

\subsection{Training Details}
\label{sec:training_details}

We trained the forward model using three Boreas-RT sequences, one each from the \texttt{Suburbs}, \texttt{Industrial}, and \texttt{Farm} routes. To maximize training-data quality, lidar submaps were constructed using ground-truth poses, and ground-truth radar poses at each azimuth timestamp were used to render the corresponding input range-image patches. Ground truth was used only for model training; all localization experiments instead used lidar-odometry-derived submaps and Dr-LiSA pose estimates.

Each training sequence was split temporally, with the first $80\%$ of radar frames used for training and the remaining $20\%$ for validation. We do not define a separate model test split because final performance is evaluated through the downstream localization task. This yields \num{7940} training frames and \num{1985} validation frames. Since each of the 400 radar azimuths constitutes an individual training sample, these correspond to approximately $3.2$ million training examples and $0.8$ million validation examples. The model was trained for 50 epochs using AdamW~\cite{loshchilov2019decoupled} with a learning rate of $5\times10^{-4}$ and a batch size of 64. The best weights were then used for all localization experiments on the 15 additional evaluation sequences described in Sec.~\ref{sec:dataset}.

\begin{table*}[!t]
    \centering
    \caption{Ablation study results on the Suburbs route. Bolding marks the best value in each column of the first block, and values that improve on the Dr-LiSA/Suburbs baseline in the remaining blocks.}
    \footnotesize{
    \begin{tabular*}{\textwidth}{@{\extracolsep{\fill}}lcccccc}
        \toprule
        \textbf{Method / Trained On}
        & \textbf{Long.} [\si{\m}]
        & \textbf{Lat.} [\si{\m}]
        & \textbf{Vert.} [\si{\m}]
        & \textbf{Roll} [$^\circ$]
        & \textbf{Pitch} [$^\circ$]
        & \textbf{Yaw} [$^\circ$] \\
        \midrule
        \textbf{Dr-LiSA} / \texttt{Suburbs}
        & \textbf{0.067} & \textbf{0.058} & \textbf{0.160} & 0.420 & 0.283 & 0.104 \\
        \textbf{Filtering: off} / \texttt{Suburbs}
        & 1.209 & 0.560 & 0.489 & 0.983 & 0.504 & 0.141 \\
        \textbf{MSE} / \texttt{Suburbs}
        & 0.080 & 0.071 & 0.252 & \textbf{0.417} & \textbf{0.253} & \textbf{0.101} \\
        \textbf{No Skip Branch} / \texttt{Suburbs}
        & 0.108 & 0.079 & 0.255 & 0.470 & 0.276 & 0.105 \\
        \midrule
        \textbf{Dr-LiSA} / \texttt{Industrial}
        & 0.075 & 0.076 & 0.656 & 1.259 & 0.440 & 0.131 \\
        \textbf{Dr-LiSA} / \texttt{Farm}
        & 0.087 & 0.083 & 0.342 & 0.546 & 0.364 & 0.152 \\
        \midrule
        \textbf{Dr-LiSA + 3DRO} {\scriptsize\cite{gentil20263dro}} / \texttt{Suburbs}
        & \textbf{0.060} & \textbf{0.054} & \textbf{0.156} & \textbf{0.306} & \textbf{0.241} & \textbf{0.088} \\
        \bottomrule
    \end{tabular*}}
    \label{tab:ablation}
\end{table*}

\begin{figure*}[!t]
    \begin{minipage}[t]{\columnwidth}
        \vspace{0pt}
        \centering
        \includegraphics[
            width=\columnwidth,
            trim=1.3cm 0.3cm 0cm 0cm,
            clip
        ]{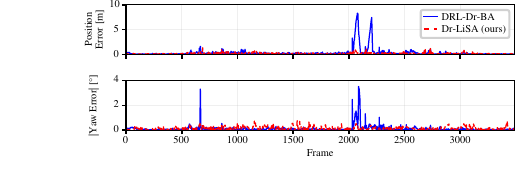}
        \caption{$SE(2)$ error comparison on a \texttt{Farm} route. Position and absolute yaw errors are shown for Dr-LiSA and DRL-Dr-BA~\cite{lisus2026drba} over the sequence.}
        \label{fig:comparison_plots}
    \end{minipage}\hfill
    \begin{minipage}[t]{\columnwidth}
        \vspace{0pt}
        \centering
        \includegraphics[
            width=\columnwidth,
            trim=1.6cm 0.3cm 0cm 0cm,
            clip
        ]{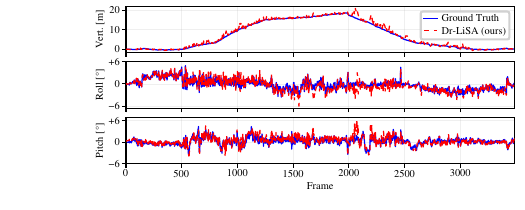}
        \caption{Dr-LiSA vertical, roll, and pitch tracking on a \texttt{Farm} sequence.}
        \label{fig:tracking_plots}
    \end{minipage}
\end{figure*}

\subsection{Localization Results}
\label{sec:results}

Dr-LiSA is compared to three baselines: a SOTA radar-lidar approach (RLT\&R)~\cite{are_we_ready_for}, a SOTA radar-radar approach (DRL-Dr-BA)~\cite{lisus2026drba}, and a SOTA lidar-lidar approach (2Fast-2Lamaa)~\cite{legentil20242fast2lamaa}. Since Dr-LiSA is the first radar-based localization approach in $SE(3)$, we include a lidar baseline as reference. To ensure a fair comparison with DRL-Dr-BA, we propagate Dr-LiSA estimates using the same $SE(2)$ radar odometry method, DRO~\cite{legentil2025dro}, rather than introducing a separate $SE(3)$ odometry backend. All approaches are evaluated on the same mapping and localization sequences.

Table~\ref{tab:localization_boreas} reports the longitudinal, lateral, vertical, roll, pitch, and yaw root-mean-square error (RMSE) across all route types. On the \texttt{Suburbs}, \texttt{Industrial}, and \texttt{Regional} routes, Dr-LiSA consistently outperforms RLT\&R across all reported metrics. We attribute this in part to the dense direct alignment formulation, which avoids relying on sparse radar point extractions that can be sensitive to radar-specific artefacts and noise. On these routes, Dr-LiSA also achieves translational accuracy comparable to DRL-Dr-BA while additionally estimating the full $SE(3)$ pose, which DRL-Dr-BA does not provide. Dr-LiSA's rotational accuracy remains somewhat lower, however. One possible explanation is that radar orientation is strongly constrained by distant scene structure, since even small angular errors produce large spatial offsets at long range. These distant returns may be well represented in radar maps, but are not always observed as reliably in lidar due to differences in sensing range, visibility, and scene coverage, potentially making rotational alignment more difficult in the radar-lidar setting.

The advantages of Dr-LiSA are most pronounced on the challenging \texttt{Farm} route, where all methods degrade and RLT\&R fails to localize through to the end of any sequence. Not only does Dr-LiSA avoid these failures, but it also substantially improves over DRL-Dr-BA in every metric that DRL-Dr-BA reports. The \texttt{Farm} route contains greater roll and pitch variation with long feature-sparse regions, conditions under which DRL-Dr-BA becomes unstable as noted in the original work. In contrast, Dr-LiSA can exploit 3D scene structure, including ground-plane returns that become informative under non-planar vehicle motion, to stabilize localization in these regions (Fig.~\ref{fig:scan_comparisons}). This results in a marked reduction in peak localization error (Fig.~\ref{fig:comparison_plots}) and demonstrates that full $SE(3)$ localization can improve $SE(2)$ robustness in feature-sparse, non-planar environments.

\subsection{Ablation Studies}
\label{sec:ablation}

Table~\ref{tab:ablation} summarizes ablations examining the effects of different training methods, cross-environment generalization, and 3D odometry. All ablations are evaluated on the \texttt{Suburbs} localization route to isolate these design choices under a common test condition.

\textit{1) Model Training:} All models in this ablation are trained on a single \texttt{Suburbs} sequence, including the Dr-LiSA baseline. Against this baseline we evaluate training on unfiltered radar measurements, a conventional MSE objective, and removal of the geometry-guided skip branch. Training on unfiltered radar measurements causes a substantial degradation in localization accuracy, particularly in translation, highlighting the importance of radar preprocessing. Replacing the co-visibility loss with a conventional MSE objective yields slightly lower rotational RMSE but noticeably worse translational performance. Removing the geometry-guided skip branch similarly degrades translation and vertical accuracy. Overall, the complete training configuration provides the best balance of performance, achieving the lowest translational errors while maintaining comparable rotational accuracy.

\textit{2) Generalizability:} To evaluate cross-environment generalization, we train separate models using only an \texttt{Industrial} sequence and only a \texttt{Farm} sequence, then evaluate both on the \texttt{Suburbs} route. The \texttt{Industrial} environment shares more structural similarity with \texttt{Suburbs}, whereas \texttt{Farm} is considerably more sparse and rural. Despite this domain shift, both models retain strong $SE(2)$ localization performance on \texttt{Suburbs}, although their $SE(3)$ accuracy degrades relative to the \texttt{Suburbs}-trained model. These results indicate that the learned lidar-to-radar mapping transfers to unseen maps and to substantially different environment types, with the largest penalty in $SE(3)$ accuracy.

\textit{3) 3D Odometry:} Finally, we replace the $SE(2)$ DRO propagation used in the main experiments with its $SE(3)$ counterpart 3DRO~\cite{gentil20263dro}. Using 3D odometry not only improves vertical, roll, and pitch accuracy, but also the planar localization metrics. This indicates that more accurate $SE(3)$ initialization benefits the subsequent direct optimization across all pose dimensions.

\subsection{Limitations}
\label{sec:limitations}

Despite enabling $SE(3)$ localization from spinning-radar measurements and achieving SOTA $SE(2)$ performance, Dr-LiSA does not yet operate in real time. On a single Lenovo ThinkPad P16 Gen 1 equipped with an Intel Core i7-12800HX CPU, NVIDIA RTX A4500 Laptop GPU, and \SI{16}{\giga\byte} of RAM, the system requires an average of \SI{2.3}{\s} per radar frame. Runtime is dominated by repeated forward-model evaluations during optimization. Future work will therefore investigate reducing model complexity and lidar range-image resolution, as well as other strategies for accelerating inference.

Although Dr-LiSA exhibits higher vertical, roll, and pitch RMSE than 2Fast-2Lamaa, it still tracks these degrees of freedom closely on challenging routes as shown in Fig.~\ref{fig:tracking_plots}. Moreover, the ablation results in Sec.~\ref{sec:ablation} show that replacing $SE(2)$ odometry with $SE(3)$ odometry improves roll, pitch, and vertical accuracy, indicating that better pose propagation can further strengthen full 6-DoF localization.

\section{Conclusion}
\label{sec:conclusion}

This paper introduces Dr-LiSA, the first method to estimate $SE(3)$ pose from 2D spinning-radar measurements. Although spinning radar produces a two-dimensional range-azimuth image, the measured intensities depend on interactions between the radar beam and the surrounding 3D scene. Across more than \SI{90}{\km} of real-world driving, Dr-LiSA outperforms the strongest prior radar-lidar baseline (RLT\&R) in $SE(2)$ and achieves planar accuracy competitive with SOTA radar-radar localization, all while tracking vertical, roll, and pitch motion. The advantages are most evident in feature-sparse, non-planar environments, where access to 3D scene structure improves localization robustness relative to $SE(2)$ approaches. While further work is required to improve runtime and non-planar accuracy, Dr-LiSA establishes a new direction for radar localization against 3D lidar maps.

\printbibliography

@string{tog =  {ACM Trans. Graph.} }

@string{accv    = {Proc. Asian Conf. Comput. Vis.} }

@string{ai      = {Artificial Intelligence} }

@string{arxiv   = {arXiv preprint} }

@string{corl    = {Proc. Conf. Robot Learn.} }

@string{crv     = {Proc. Conf. Comput. Robot Vis.} }

@string{eccv    = {Proc. Eur. Conf. Comput. Vis.} }

@string{iccv    = {Proc. IEEE Int. Conf. Comput. Vis.} }

@string{icra    = {Proc. IEEE Int. Conf. Robot. Autom.} }

@string{ijrr    = {Int. J. Robot. Res.} }

@string{iros    = {Proc. IEEE/RSJ Int. Conf. Intell. Robots Syst.} }

@string{nipsjournal={Adv. Neural Inf. Process. Syst.} }

@string{ral     = {IEEE Robot. Autom. Lett.} }

@string{rss     = {Proc. Robot.: Sci. Syst.} }

@string{sensors = {IEEE Sensors Journal} }

@string{taes     = {IEEE Trans. Aerosp. Electron. Syst.} }

@string{cvpr = {Proc. IEEE/CVF Conf. Comput. Vis. Pattern Recognit.}}

@string{qce = {Proc. IEEE Int. Conf. Quantum Comput. Eng.}}

@string{iclr = {Proc. Int. Conf. Learn. Represent.}}

@Article{s26165316,
AUTHOR = {Viktor, Patrik},
TITLE = {From Modality Performance to Graceful Degradation: A PRISMA 2020 Systematic Review of Sensor Architectures for Autonomous Vehicles},
JOURNAL = {Sensors},
VOLUME = {26},
YEAR = {2026},
NUMBER = {16},
ARTICLE-NUMBER = {5316},
URL = {https://www.mdpi.com/1424-8220/26/16/5316},
PubMedID = {42655624},
ISSN = {1424-8220},
DOI = {10.3390/s26165316}
}

@article{reid2019localization,
  title   = {{Localization Requirements for Autonomous Vehicles}},
  author  = {Reid, T. G. and Houts, S. E. and Cammarata, R. and others},
  journal = {SAE Int. J. Connected Automated Veh.},
  volume  = {2},
  number  = {3},
  pages   = {173--190},
  year    = {2019}
}

@inproceedings{wang2020l2rgan,
  title     = {{L2R GAN: LiDAR-to-Radar Translation}},
  author    = {Wang, Leichen and Goldluecke, Bastian and Anklam, Carsten},
  booktitle = accv,
  year      = {2020}
}

@article{abu2023radar,
  title={{Radar Odometry for Autonomous Ground Vehicles: A Survey of Methods and Datasets}},
  author={Abu-Alrub, Nader J and Rawashdeh, Nathir A},
  journal={IEEE Trans. Intell. Veh.},
  volume={9},
  number={3},
  pages={4275--4291},
  year={2023},
  publisher={IEEE}
}

@inproceedings{Park_Kim_Kim_2019,
  title={{Radar Localization and Mapping for Indoor Disaster Environments via Multi-modal Registration to Prior LiDAR Map}},
  author={Park, Yeong Sang and Kim, Joowan and Kim, Ayoung},
  booktitle=iros,
  pages={1307--1314},
  year={2019}
}

@inproceedings{radar_on_lidar,
  title={{Radar-on-lidar: Metric Radar Localization on Prior Lidar Maps}},
  author={Yin, Huan and Wang, Yue and Tang, Li and Xiong, Rong},
  booktitle={Proc. IEEE Int. Conf. Real-Time Comput. Robot.},
  pages={1--7},
  year={2020},}

@article{RaLL,
  title={{RaLL: End-to-end Radar Localization on Lidar Map Using Differentiable Measurement Model}},
  author={Yin, Huan and Chen, Runjian and Wang, Yue and Xiong, Rong},
  journal={IEEE Trans. Intell. Transp. Syst.},
  volume={23},
  number={7},
  pages={6737--6750},
  year={2021},
  publisher={IEEE}
}

@inproceedings{RoLM,
  title={{RoLM: Radar on LiDAR Map Localization}},
  author={Ma, Yukai and Zhao, Xiangrui and Li, Han and Gu, Yaqing and Lang, Xiaolei and Liu, Yong},
  booktitle=icra,
  pages={3976--3982},
  year={2023}
}

@article{Yin_Xu_Wang_Xiong_2021,
  title={{Radar-to-lidar: Heterogeneous Place Recognition via Joint Learning}},
  author={Yin, Huan and Xu, Xuecheng and Wang, Yue and Xiong, Rong},
  journal={Front. Robot. AI},
  volume={8},
  eid={661199},
  year={2021},
  publisher={Frontiers Media SA}
}

@article{are_we_ready_for,
  title={{Are We Ready for Radar to Replace Lidar in All-weather Mapping and Localization?}},
  author={Burnett, Keenan and Wu, Yuchen and Yoon, David J and Schoellig, Angela P and Barfoot, Timothy D},
  journal=ral,
  volume={7},
  number={4},
  pages={10328--10335},
  year={2022},
  publisher={IEEE}
}

@inproceedings{lisus2025pointing,
  author={Lisus, Daniil and Laconte, Johann and Burnett, Keenan and Zhang, Ziyu and Barfoot, Timothy D.},
  title={{Pointing the Way: Refining Radar-Lidar Localization Using Learned ICP Weights}},
  booktitle=crv,
  year={2025}
}

@article{a_new_wave_radar,
  title={{A New Wave in Robotics: Survey on Recent mmWave Radar Applications in Robotics}},
  author={Harlow, Kyle and Jang, Hyesu and Barfoot, Timothy D and Kim, Ayoung and Heckman, Christoffer},
  journal={IEEE Trans. Robot.},
  year={2024},
  publisher={IEEE}
}

@article{venon2022millimeter,
  title={{Millimeter Wave FMCW Radars for Perception, Recognition and Localization in Automotive Applications: A Survey}},
  author={Venon, Arthur and Dupuis, Yohan and Vasseur, Pascal and Merriaux, Pierre},
  journal={IEEE Trans. Intell. Veh.},
  volume={7},
  number={3},
  pages={533--555},
  year={2022},
  publisher={IEEE}
}

@inproceedings{under_the_radar,
  author = {Dan Barnes and Ingmar Posner},
  title = {{Under the Radar: Learning to Predict Robust Keypoints for Odometry Estimation and Metric Localisation in Radar}},
  booktitle=icra,
  year = {2020}
}

@inproceedings{barnes2019masking,
  title={{Masking by Moving: Learning Distraction-Free Radar Odometry from Pose Information}},
  author={Barnes, Dan and Weston, Rob and Posner, Ingmar},
  booktitle=corl,
  pages={303--316},
  year={2020}
}

@INPROCEEDINGS{burnett_rss21,
    title={{Radar Odometry Combining Probabilistic Estimation and Unsupervised Feature Learning}},
    author={Burnett, Keenan and Yoon, David J and Schoellig, Angela P and Barfoot, Timothy D},
    booktitle=rss,
    year={2021}
}

@inproceedings{Cen_filtering,
  title={{Precise Ego-Motion Estimation With Millimeter-Wave Radar Under Diverse and Challenging Conditions}},
  author={Cen, Sarah H and Newman, Paul},
  booktitle=icra,
  pages={6045--6052},
  year={2018}
}

@article{rohling1983radar,
  title={{Radar CFAR Thresholding in Clutter and Multiple Target Situations}},
  author={Rohling, Hermann},
  journal=taes,
  number={4},
  pages={608--621},
  year={1983},
  publisher={IEEE}
}

@INPROCEEDINGS{mcl,
  author={Dellaert, F. and Fox, D. and Burgard, W. and Thrun, S.},
  booktitle=icra,
  title={{Monte Carlo Localization for Mobile Robots}},
  year={1999},
  volume={2},
  pages={1322--1328},
  doi={10.1109/ROBOT.1999.772544}}

@inproceedings{unet,
  title={{U-Net: Convolutional Networks for Biomedical Image Segmentation}},
  author={Ronneberger, Olaf and Fischer, Philipp and Brox, Thomas},
  booktitle={Proc. Int. Conf. Med. Image Comput. Comput.-Assist. Interv.},
  pages={234--241},
  year={2015},}

@article{VTR,
  title={{Visual Teach and Repeat for Long-Range Rover Autonomy}},
  author={Furgale, Paul and Barfoot, Timothy D},
  journal={J. Field Robot.},
  volume={27},
  number={5},
  pages={534--560},
  year={2010},
  publisher={Wiley Online Library}
}

@ARTICLE{2021_Burnett,
  author={Burnett, Keenan and Schoellig, Angela P. and Barfoot, Timothy D.},
  journal=ral,
  title={{Do We Need to Compensate for Motion Distortion and Doppler Effects in Spinning Radar Navigation?}},
  year={2021},
  volume={6},
  number={2},
  pages={771-778},
  doi={10.1109/LRA.2021.3052439}
}

@inproceedings{huang2023nksr,
  title     = {{Neural Kernel Surface Reconstruction}},
  author    = {Huang, Jiahui and Gojcic, Zan and Atzmon, Matan and Litany, Or and Fidler, Sanja and Williams, Francis},
  booktitle = cvpr,
  pages     = {4369--4379},
  year      = {2023}
}

@article{parker2010optix,
  title={{OptiX: A General Purpose Ray Tracing Engine}},
  author={Parker, Steven G and Bigler, James and Dietrich, Andreas and Friedrich, Heiko and Hoberock, Jared and Luebke, David and McAllister, David and McGuire, Morgan and Morley, Keith and Robison, Austin and others},
  journal=tog,
  volume={29},
  number={4},
  pages={1--13},
  year={2010},
  publisher={ACM New York, NY, USA}
}

@inproceedings{ranftl2021vision,
  title={{Vision Transformers for Dense Prediction}},
  author={Ranftl, Ren{\'e} and Bochkovskiy, Alexey and Koltun, Vladlen},
  booktitle=iccv,
  pages={12159--12168},
  year={2021},}

@inproceedings{dosovitskiy2020image,
  title={{An Image is Worth 16x16 Words: Transformers for Image Recognition at Scale}},
  author={Dosovitskiy, Alexey and Beyer, Lucas and Kolesnikov, Alexander and Weissenborn, Dirk and Zhai, Xiaohua and Unterthiner, Thomas and Dehghani, Mostafa and Minderer, Matthias and Heigold, Georg and Gelly, Sylvain and others},
  booktitle=iclr,
  year={2021}
}

@inproceedings{zheng2021rethinking,
  title={{Rethinking Semantic Segmentation from a Sequence-to-Sequence Perspective with Transformers}},
  author={Zheng, Sixiao and Lu, Jiachen and Zhao, Hengshuang and Zhu, Xiatian and Luo, Zekun and Wang, Yabiao and Fu, Yanwei and Feng, Jianfeng and Xiang, Tao and Torr, Philip HS and others},
  booktitle=cvpr,
  pages={6877--6886},
  year={2021},}

@inproceedings{carion2020end,
  title={{End-to-End Object Detection with Transformers}},
  author={Carion, Nicolas and Massa, Francisco and Synnaeve, Gabriel and Usunier, Nicolas and Kirillov, Alexander and Zagoruyko, Sergey},
  booktitle=eccv,
  pages={213--229},
  year={2020}
}

@article{mikolov2013distributed,
  title={{Distributed Representations of Words and Phrases and Their Compositionality}},
  author={Mikolov, Tomas and Sutskever, Ilya and Chen, Kai and Corrado, Greg S and Dean, Jeff},
  journal=nipsjournal,
  volume={26},
  year={2013}
}

@ARTICLE{lisus2025doppler,
  author={Lisus, Daniil and Burnett, Keenan and Yoon, David J. and Poulton, Richard and Marshall, John and Barfoot, Timothy D.},
  journal=ral,
  title={{Are Doppler Velocity Measurements Useful for Spinning Radar Odometry?}},
  year={2025},
  volume={10},
  number={1},
  pages={224-231},
  doi={10.1109/LRA.2024.3505821}
}

@book{barfoot2024state,
  title={{State Estimation for Robotics}},
  author={Barfoot, Timothy D},
  year={2024},
  publisher={Cambridge University Press}
}

@article{pytorch,
  title={{PyTorch: An Imperative Style, High-performance Deep Learning Library}},
  author={Paszke, Adam and Gross, Sam and Massa, Francisco and Lerer, Adam and Bradbury, James and Chanan, Gregory and Killeen, Trevor and Lin, Zeming and Gimelshein, Natalia and Antiga, Luca and others},
  journal=nipsjournal,
  volume={32},
  year={2019}
}

@inproceedings{lavrijsen2020classical,
  author    = {Lavrijsen, Wim and Tudor, Ana and M{\"u}ller, Juliane and Iancu, Costin and de Jong, Wibe},
  title     = {{Classical Optimizers for Noisy Intermediate-Scale Quantum Devices}},
  booktitle = qce,
  year      = {2020},
  pages     = {267--277},
  doi       = {10.1109/QCE49297.2020.00041}
}

@book{kelley2011implicit,
  title={{Implicit Filtering}},
  author={Kelley, Carl T},
  year={2011},
  publisher={SIAM}
}

@inproceedings{loshchilov2019decoupled,
  title     = {{Decoupled Weight Decay Regularization}},
  author    = {Loshchilov, Ilya and Hutter, Frank},
  booktitle = iclr,
  year      = {2019}
}

@inproceedings{lisus2026drba,
  author={Lisus, Daniil and {Le Gentil}, Cedric and Barfoot, Timothy D.},
  title={{Dr-BA: Separable Optimization for Direct Radar Bundle Adjustment \& Localization}},
  booktitle=rss,
  year={2026}
}

@inproceedings{legentil2025dro,
  title={{DRO: Doppler-Aware Direct Radar Odometry}},
  author={{Le Gentil}, Cedric and Brizi, Leonardo and Lisus, Daniil and Qiao, Xinyuan and Grisetti, Giorgio and Barfoot, Timothy D.},
  booktitle=rss,
  year={2025}
}

@article{lisus_brrt26,
  author={Daniil Lisus and Katya M. Papais and Cedric {Le Gentil} and Elliot Preston-Krebs and Andrew Lambert and Keith Y. K. Leung and Timothy D. Barfoot},
  title={{Boreas Road Trip: A Multi-Sensor Autonomous Driving Dataset on Challenging Roads}},
  journal=ijrr,
  year={2026},
  note={to appear}
}

@article{legentil20242fast2lamaa,
  title={{2Fast-2Lamaa: Large-Scale Lidar-Inertial Localization and Mapping with Continuous Distance Fields}},
  author={{Le Gentil}, Cedric and Falque, Raphael and Lisus, Daniil and Barfoot, Timothy D.},
  journal=ijrr,
  year={2026},
  doi={10.1177/02783649261451003}
}

@misc{gentil20263dro,
  title={{3DRO: Lidar-Level SE(3) Direct Radar Odometry Using a 2D Imaging Radar and a Gyroscope}},
  author={{Le Gentil}, Cedric and Lisus, Daniil and Barfoot, Timothy D.},
  year={2026},
  eprint={2604.12027},
  archivePrefix={arXiv},
  primaryClass={cs.RO}
}

\end{document}